\documentclass[runningheads]{llncs}
\usepackage [pagewise]{lineno}
\usepackage{tikz}
\usepackage{amsmath}
\usepackage{amssymb}
\usepackage{booktabs}
\usepackage{multirow}
\usepackage{pifont}
\usepackage{subcaption}

\newcommand{\npar}[1]{\noindent\textbf{#1}}

\usepackage{graphicx}
\usepackage{amssymb}
\begin{document}
\title{Multi-Modal interpretable automatic video captioning}
\author{Antoine Hanna-Asaad\inst{1,2} \and
Decky Aspandi\inst{2} \and
Titus Zaharia\inst{2}}
\authorrunning{Antoine et al.}
\institute{École Supérieure des Techniques Aéronautiques et de Construction Automobile (ESTACA), France 
\email{antoine.hannaasaad@estaca.eu}\\
\and
Department ARTEMIS, Télécom SudParis, France \\
}

%\email{\{name,surname\}@telecom-sudparis.eu}
\maketitle              % typeset the header of the contribution
\begin{abstract}
Video captioning aims to describe video contents using natural language format that involves understanding and interpreting scenes, actions and events that occurs simultaneously on the view. Current approaches have mainly concentrated on visual cues, often neglecting the rich information available from other important modality of audio information, including their inter-dependencies. In this work, we introduce a novel video captioning method trained with multi-modal contrastive loss that emphasizes both multi-modal integration and interpretability. Our approach is designed to capture the dependency between these modalities, resulting in more accurate, thus pertinent captions. Furthermore, we highlight the importance of interpretability, employing multiple attention mechanisms that provide explanation into the model’s decision-making process. Our experimental results demonstrate that our proposed method performs favorably against the state-of the-art models on commonly used benchmark datasets of MSR-VTT and VATEX.

\end{abstract}
%
%
%
%\linenumbers

\section{Introduction}

Video captioning has the goal of automatically generating sentence to describe the video content \cite{venugopalan2015sequence,guadarrama2013youtube2text}. This area of research  has emerged as an important area with multiple applications ranging from content indexing and retrieval \cite{ma2015multimodal,song2018quantization} to assist individuals with visual impairments \cite{voykinska2016blind} . However, complex nature of the video presents a significant challenge for captioning tasks. Compared to image captioning \cite{chen2017sca,vinyals2016show,lu2017knowing}, videos consists of temporal sequences where multiple actions and events occur in the same time or in rapid succession. Additionally, audio cues, like as speech, music, and environmental sounds, introduce an extra layer of complexity that need to be managed to generate coherent caption.

Despite making substantial progress, existing methods in this task is still considered inadequate in capturing the local and global representation. Various works often relies only on raw-pixels and one modality only \cite{wang2018reconstruction,nadeem2023sem,ye2022hierarchical,zhou2018end} and often neglect other modalities such as audio, speech or audio-caption. To address this, recent multi-modal approaches \cite{chen2023valor,tian2019audio,iashin2020multi} integrate various modalities, however,  work to relate such modalities is still lacking. This can be problematic, given that there exist depedency between these modalities on the video (i.e. video is mainly composed of both Auditory and Visual information that is interconnected). This is also compounded by a significant challenge lies in the interpretability of such complex systems. Understanding the reasoning behind a model's caption generation is crucial for enhancing trust in these models. Thus our works will serve as the first mutli-modal work combining audio-caption and interpretability for video-captioning, that also relates these modalities on the same time.

%*Futhermore, current method doesn't exploit the rich relations between modality with the use of contrastive losses.* %

%NAME of Model%

In this work, we propose a novel video captioning framework that leverages neural architectures and multi-modal fusion techniques to address aforementioned challenges. Our approach integrates visual and auditory information through a unified encoder-decoder model, utilizing attention mechanisms to focus on relevant features across different modalities. We also incorporate interpretability techniques to provide insights into the model's decision-making process. We evaluate our model on benchmark datasets. Hence, the contributions of this work are as follow:
%, demonstrating significant improvements in caption quality, coherence, and interpretability compared to existing methods.

\begin{enumerate}
    \item We introduce a multi-modal fusion strategy that effectively combines visual and audito features, enhancing the model's ability to generate higher quality captions.
    \item We incorporate interpretability techniques to analyze and visualize the model's attention, providing  insights into its decision-making process.
    \item We show our competitive results on both well established and largest video captioning datasets of MSR-VTT and VATEX.
\end{enumerate}

The remainder of this paper is structured as follows: Sec ~\ref{sec:related_works} reviews related work in video captioning. Sec~\ref{sec:Methodology} details our proposed methodology, including the model architecture and training process. Sec~\ref{sec:results} presents experimental results and analysis. Finally, Section ~\ref{sec:Conclusion} concludes the paper and outlines potential directions for future research.

\section{Related Works}
\label{sec:related_works}

\npar{Video Captioning} attempts to produce a natural language sentence that describes the action, within a given video observation \cite{abdar2023review}. The first attempt on this task can be traced to the rule-based method \cite{barbu2012video} that introduced the subject, verb, and object concepts within the caption sentencese. In their method, an encoder-decoder framework was used to extract and refine visual features, and a decoder is added to operate on top of those features to generate captions. Furthermore, several other approaches started to utilize a pre-trained 2D and 3D-CNN \cite{yang2023vid2seq,aafaq2019spatio,Perez-Martin_2021_WACV} as their visual backbone to benefit from the learned features during training with large dataset. Subsequently, they used RNNs or LSTMs methods for caption generation \cite{venugopalan2014translating,gan2017semantic} to model the sequence, natural to these video data. Another method such as \cite{ye2022hierarchical}, proposes the hierarchical multi-modal attention that selectively attends to a certain modality when generating descriptions. However, when compared to the attention mechanism~\cite{vaswani2017attention}, those methods typically perform worse because of the limitations of RNNs and LSTMs in fully capturing long-term dependencies efficiently. %readp,% to capture long-term dependencies.

In recent years, we have seen a rise in transformer based models \cite{vaswani2017attention} for video captioning due to their ability to capture long range context efficiently, due to possibility of parallelisation of attention mechanism over time. Such model of \cite{wang2022git} uses a simplified architecture similar to \cite{radford2019language} with the only difference being that the model is also conditioned on the visual features. Another method such as SwinBERT \cite{lin2022swinbert}, further proposes a sparse transformer architecture that reduces the computations needs. Finally a well known method of BLIP \cite{li2022blip} is also built upon a transformer architecture consisting of an image encoder and a text decoder. It employs a multi-task pre-training approach which uses contrastive learning to aid in different downstream tasks, including captioning.

%However, due to the fact that they align the visual informations with the ground truth video caption (accesble during training) this can lead to suboptimal result during inference (given that the video caption is not present)%

\subsubsection{Multi-Modal Captioning}
Recent advancements in the field artificial intelligence have shown that using multi-modal information for the input can greatly improve the quality of the their prediction~\cite{aspandi2022audio,zhang2023eyeGaze}. One method example for captioning task is ActBERT \cite{zhu2020actbert}  which uses a BERT-style \cite{devlin2018bert} with mask-modelling objective applied to both video and ASR (Automatic Speech Recognition) as input in their pipeline. This method indeed achieves a superior accuracy compared to a single modality, however its limitation is from the use of ASR that has high innacuracy in the transcription process which can affect the result. In recent years nevertheless, we have seen a convergence toward a unified model that also further exploit the attention mechanisms to combine the modalities. Some example include \cite{chen2024vast,chen2023valor,luo2020univl}, which encode modalities by using different pre-training objective such as masked language modeling \cite{sun2019videobert,zhou2020unified},  masked-frame-modeling \cite{li2022blip} or video-text matching \cite{li2020hero,sun2019learning,Miech_2020_CVPR}, that lead to current state-of-the-art result. %no contrastive loss.

\subsubsection{Explainability}

Explainability in captioning task is a considered as an important area of research with the aim of improving the understanding of inner workings mechanism from the existing captioning methods. One of the primary methods for achieving explainability in this task is through attention mechanisms, that it allows to trace the models focus during prediction. Specifically, these mechanisms enable the model to focus on specific parts of an input modality (e.g. image) when generating each word of the caption, providing insights into which spatial regions of the modality are considered important. Xu et al. \cite{xu2015show} introduced an attention-based model for image captioning that generates attention maps, highlighting relevant areas in the image input for each generated word, thus making the model's decision process more interpretable. Lastly, another notable technique that allows the explanation on the visual domains is Grad-CAM \cite{selvaraju2020grad} (Gradient-weighted Class Activation Mapping) and Integrated Gradients \cite{sundararajan2017axiomatic} which can be used to produce saliency maps. These maps show the gradient of the output with respect to the input image that helps in understanding which regions that contribute most to the generated caption.

%In a similar fashion, Kobayashi et al. \cite{kobayashi2020attention} analyze the attention mechanism of the existing methods with vector norms operation, further highlighing more important information. 

\section{Methodology}
\label{sec:Methodology}

\begin{figure*}[!t]
\centering
    \includegraphics[scale=0.20]{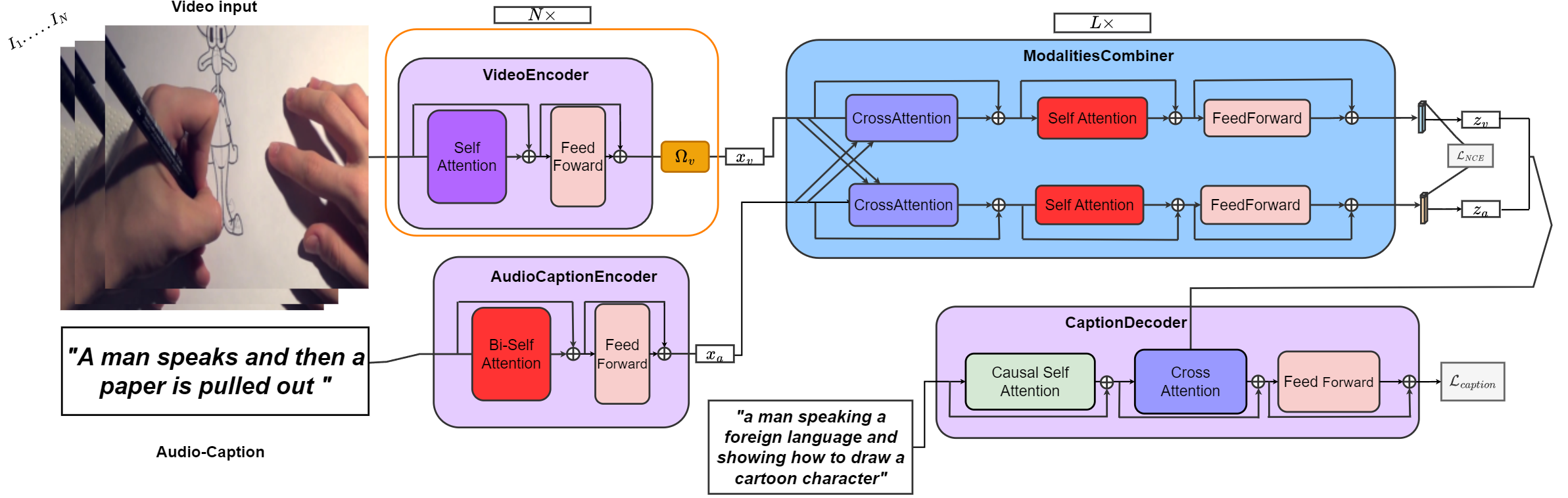}
    \caption[]{Overview of our proposed methods that involves both Video and Respective Audio Captions to generate a Video Caption.} 
    \label{fig:model}
\end{figure*}

Figure~\ref{fig:model} provides an overview of our Multi-modal Interpretable Captioner (MICap) that includes several key components: the input encoders $f(.)$ and $g(.)$, the cross modal fusion block $\psi (.)$ and the caption decoder $h(.)$. This approach is inspired from the previous works ~\cite{li2022blip,wang2022git,lin2022swinbert} with the key modifications involving our novel combiner block and the involvement of additional modality of the audio captions.

The input of our pipelines consists of sequence of images that provides the visual information, and the audio based captions that gives the information from audio points of views. These both modalities are first fed into Modalities Encoders (that consists of Video and Audio encoders) to extract their features. Then, these features are fed into the Modalities Combiner for the alignment and fusion operation, that yields both aligned and combined features. These features then fed into the CaptionDecoder, that further merged with first token of the input text to generate the final video caption. The details of each process will be described on the following section.

% Our model is depicted in Fig ~\ref{fig:model}, the process starts with the input of the video and audio caption, which is then received by our cross-modal block before going into the decoder.

%As shown in Fig. ~\ref{fig:model}, we follow the common practice in video captioning~\cite{li2022blip,wang2022git,lin2022swinbert}, and use an encoder-decoder framework including four main modules : a video encoder $f(.)$, an audio caption encoder $g(.)$ , a cross modal encoder $\psi (.)$ and a caption decoder $h(.)$.

% In this section, we explain in detail our methods for audio-visual
% video captioning. In Sec. ~\ref{subsec:problem}, we first define our objective and constraint. In Sec.~\ref{subsec:overview}, we first present an overview of our method. In Sec.~\ref{subsec:crossfusion} we describe our approach to encode the different modalities during pre-training.  In Sec.~\ref{subsec:caption},
% we detail our caption generation approach, which leverages both audio and visual features to produce comprehensive and contextually rich captions.

% \subsection{Input Encoders}
% \label{subsec :features}

\subsection{Modalities Encoders}
\label{subsec:problem}
Given a video, $\mathbf{\mathcal{V}} =\mathbf{ \{ I_{1},I_{2},...,I_{N} \}} $ where $N$ is the number of frames, and each $I_i \in \mathbb{R}^{h \times w \times c}$ and a set of audio caption $\mathcal{A}$, we obtain the generated caption $\hat{y}$ as follow:

\begin{equation}
 \mathbf{\hat{y}=h_{decoder}(f_{v-enc}(\mathcal{V}), g_{ac-enc}(\mathcal{A}))}
\end{equation}

we first extract corresponding visual features ($x_v$) using the $f_v$ for every frames with a pre-trained CLIP-ViT-B/16 \cite{radford2021learning}. Then, we further consider temporal information between each frames using set of learnable embeddings $\Omega_v$ associated to each frame. For audio encoder $g_{ac}$, we use the BERT model \cite{devlin2018bert} to process the generated audio captions (from MS-CLAP \cite{elizalde2023clap}) to extract the audio feature $x_a$. These operations can be seen below: 

\begin{equation}  
x_v=CLIP({v}_i) + \Omega_{v_i}
=f(\mathcal{V}) \in \mathbb{R}^{T\times N \times D}
\end{equation}

\begin{equation}
x_a=g(\mathcal{A}) \in \mathbb{R}^{S \times D}
rg\end{equation}

where D is the hidden dimension that is identical for both $f(.)$ and $g(.)$.  $T$ corresponds to the sequence length  of the video and $S$ the sequence length of the given audio caption. Here, the visual-encoder, audio-encoder and text-decoder are initialized from the BLIP \cite{li2022blip} weights.

\subsection{Modalities Combiner}
\label{subsec:crossfusion}
Inspired by other video understanding methods of \cite{tan2019lxmert,su2019vl}, we fuse both of video and audio modality with the use of a co-attention transformer block \cite{vaswani2017attention} that consists of two internal encoders in paralel. This encoder processes different modality, e.g. image and audio, thus resulting in correlations between modalities (video-audio and audio-video). For both the video-audio branch and audio-video branch, we have : 
\begin{equation}
z_v = Transformer(x_v,x_a),  z_a =Transformer(x_a,x_v)
\end{equation}
\label{eq:transformer}

This operation can be repeated L times. In our case, the transformer layer is composed of :

$$ Transformer(X,Z) = FFB(MHA(X, Z, Z))$$

where MHA is multiple heads of attention functions and a Feed-Forward Block (FFB). MHA is the correlation functions that consists of several elements including $N_q$ query vector of dimension $d$, $Q \in \mathbb{R}^{N \times D}$, coming from X
and $N_v$ key-value pairs, $ K \in \mathbb{R}^{N_v \times D}, V \in \mathbb{R}^{N_v \times D}$ coming from Z . Given these, attention is defined as the scale-dot product of queries $Q$ and keys $K$ that further aggregated by the value $V$: 

\begin{equation}
    Att(Q,K,V)=Softmax(\frac{Q \cdot K^T}{\sqrt{d}})V
\end{equation}

Then, we aggregate the features to be used for further caption generation process: 
\begin{equation}
    z_{va}=norm( [z_v;z_a])
\end{equation}
where  $norm (. )$  denote the Layer Normalisation operation and [;] denote the concatenation operation.

% Before going into details into our fusion strategy, we will briefly introduce some key concepts of transformer. 

% \subsubsection{Fusion} 
% \label{subsec:Fusion}
% We then feed $x_a$ and $x_v$ into our two-way Cross-Modal Transformer 

%We denote its output as  :

% \subsubsection{Background} 
% \label{subsec:background}

\subsection{Caption Decoder}
\label{subsec:caption}

Given the previously correlated features $z_{va}$, then we employ an auto-regressive decoder $h(.)$ and we use the commonly used loss objective of minimizing the negative log-likelihood to generate the target video caption. During training, the decoder operates on the sequences of tokens up to t-1 tokens, denoted as $x_{<t}$. During testing, we use the start token $[CLS]$ to initiate the auto-regressive generation of caption. The aggregated features of both $z_v$ and $z_a$ are then injected during generation through a cross attention mechanism. 

\begin{equation}
\hat{y}  = h( x_{<t}, z_{va})
\end{equation}

\begin{equation}
    \hat{y}  = \text{argmax(softmax}(W \cdot Transformer ( x_{<t},z_{va}) ))
\end{equation}

where the Transformer block is defined in eq.~\ref{eq:transformer} and W is projection matrix used to project the result back into our vocabulary size. Here, we initialize our CaptionDecoder model from a trained BERT model \cite{devlin2018bert} and we use causal masking to prevent information leakage during training. Finally, we use beam-search with a beam width of five for the caption generation during inference.

%SEED,y_hat,ouput

\subsection{Loss Functions and Models Training}

To optimise our predictions, we employ two main losses, namely the caption loss and the contrastive loss. The caption loss concerns the quality of the predicted captions to the ground truth caption and defined as : 
\begin{equation}
\mathcal{L}_{caption}= -\mathbb{E}_{x \sim p_{data}} \log p_{\theta} \big(x_t| x_{<t}, z_{va} \big)\\
\end{equation}

where $p_{data}$ corresponds to our data distribution, $ p_{\theta}$ represent our decoder,  $x_t$ denote the ground truth caption at step t, $x_{<t}$ denote the previous t-1 tokens, and $\theta$ correspond to the trainable parameters.

We then use contrastive loss to extract more relevant features by evaluating their value in embedding space, subject to the contrasting criteria. This criteria assumes that extracted visual and audio features that belongs to the same video to have similar representation in embedding space compared to ones that come from different video, and vice versa. To do this, we perform the alignment operation \cite{gutmann2010noise} that is common in another relevant task (e.g. video retrieval \cite{akbari2021vatt,wu2023cap4video} ) during optimisation. This operation effectively aligns the features frames and audio caption from the same video to be as close as possible. This operations starts by performing pooling operation on the output of the cross-module for the video and audio

\begin{equation}
 c_v=Pool_{video}(z_v),  c_a=Pool_{audio}(z_a)
\end{equation}

where "pool" refers to obtaining the hidden state corresponding to the first token followed by a Tanh activation on the projected token. Then for each video and text, we compute the loss by obtaining normalized-softmax values from text-to-video and video-to-text cosine similarity :

\begin{equation}
q^{a2v}_l(a) \frac{\exp({s(a,v_l)/\tau})}{ \sum_{l=1}^L \exp(s(a,v_l)/\tau)} ,
 q^{v2a}_l(v)=\frac{\exp(s(v,a_l)/\tau)}{ \sum_{l=1}^L \exp(s(v,a_l)/\tau)} \\
\end{equation}

with \begin{equation}
    s(a,v) = \frac{c_a^T \cdot c_v}{||c_a ||_2 || c_v||_2} \in [-1,1] 
\end{equation} 

where s() denotes the cosine similarity between audio caption and video modality, and $\tau$ design the temperature. Then, letting $y^{a2v}(v)$ and $y^{v2a}(a)$ to denote the ground-truth one-hot similarity, where negative pairs have a probability of 0 and the positive pair has a probability of 1,  the video to text contrastive loss is defined as the cross-entropy between q and y: 

\begin{equation}
  \mathcal{L}_{nce}= \frac{1}{2} \mathbb{E}_{(a,v) \sim \mathcal{B}} [H(y^{v2a}(v),q^{v2a}(v)) + H(y^{a2v}(a),q^{a2v}(a)) ]
\end{equation}

% \subsection{Model Training}
% \label{subsec:Training}

Lastly, we train the model with the combine loss of $\mathcal{L}$ that includes both previously defined losses: 
\begin{equation}
    \mathcal{L}=  \mathcal{L}_{nce} +\mathcal{L}_{caption} 
\end{equation}

\subsection{Implementation Details}
% $5 \times
For pre-processing stage, we sample the frames from each video clip using ffmpeg library \footnote{https://github.com/FFmpeg/FFmpeg.git} and we resize and crop the frames into an image of size of $ 224 \times 224 \times 3$, while the corresponding, extracted audio-caption is padded with S value of 67 long. We train all models using a single NVIDIA-3080Ti for approximately one to five days for MSR-VTT and VATEX repectively. We use a learning rate of 1e-5 for the text decoder and the modalities combiner, a learning rate of 5e-5 for the input encoders, with a $\tau$ value (for the contrastive loss) set to 0.07. We use gradient accumulation and half precision to speed up the training process and reduce the memory overhead of our model simultaneously. The implementation of our methods can be found on the following repository\footnote{https://github.com/toto-a/MICap}.

% the training process. Lastly, the implementation code of our method can be found in our code repository \footnote{}.

%, while still preserving the model's overall capabilities. 

% \subsubsection{Modality Balancing (Video-AC Alignement)}.

% In order to train the video-text alignment network, we use the NCE loss  which has been
% adopted for such video-text alignment problem \\

% The Loss function is defined as follows :

\section{Experiment Results}
\label{sec:results}

\subsection{Datasets and Experiment Settings}

Two commonly used video captioning datasets are considered for the evaluations: 
\begin{enumerate}
     \item \npar{MSRVTT}\cite{xu2016msr} dataset which contains $10k$ open-domain video clips for video captioning. Each video clip is annotated with $20$ captions. To comply with Youtube copyright regulations, we are left with $6,8k$ video clips. 
     %\item \npar{MSVD}\cite{xu2016msr} dataset which contains $2k$ open-domain video clips for video captioning. 
    \item \npar{VATEX} \cite{wang2019vatex} dataset is multilingual, large, linguistically complex, and diverse dataset in terms of both video and natural language descriptions. It is composed of over $41k$ videos. We use the standard split defined in \cite{wang2019vatex}
\end{enumerate}

To evaluate the quality of the predictions, we use four quantitative metrics of BLEU@4 \cite{papineni2002bleu}, METEOR \cite{banerjee2005meteor}, ROUGE-L \cite{lin2004rouge}, and CIDEr \cite{vedantam2015cider}. Each metric provides a distinct insight into the quality of the generated captions. BLEU@4 measures sentence fluency, METEOR evaluates semantic relevance, ROUGE-L assesses the sequence of words, and CIDEr gauges how well the caption captures key details. Evaluating these different metrics allows for a more comprehensive assessment of each method's performance.

\subsection{The impact of Fusion and Contrastive Loss}

Here we present the comparison against five versions of our method to highlight the impacts of each of our approaches: 

\begin{itemize}
    \item \textbf{BLIP}~\cite{li2022blip} is external method with similar network arrangement to ours that operates mainly with the use of visual modality. 
    \item \textbf{Audio Raw} is the results which directly compare the audio-caption input by \cite{elizalde2023clap} against the ground truth test caption.
    \item \textbf{Audio-Based} is our method that uses audio-caption as input and trained with both AudioCaption encoder and CaptionDecoder.
    \item \textbf{Vision-Based} is our method that uses the videos as input and trained with both the VideoEncoder encoder and CaptionDecoder.
    \item \textbf{Fusion} is our model variant that uses video and audio modality as the input and is trained using only the captioning loss.
    \item \textbf{MICap} is our full model that uses video and audio modalities, including our proposed combined loss.
\end{itemize}

% \begin{table}[h!]
% \centering
% \caption{Table result for different models versions on MSVD and MSR-VTT}
% \label{tab:merged_result}
% \begin{tabular}{|l|l|l|l|l|l|l|l|l|l|} 
% \hline
% \multirow{2}{*}{No.} & \multirow{2}{*}{Models} & \multicolumn{4}{|c|}{MSVD} & \multicolumn{4}{|c|}{MSR-VTT} \\ \hline  \cline{3-10}
%  &  & BLEU@4& ROUGE-L & METEOR & CIDEr & BLEU-4 & ROUGE-L & METEOR & CIDEr \\ \hline
% 1. & BLIP \cite{li2022blip} & 30.5 & 56.7 & 30.7 & 61.4 & 18.2 & 44.2 & 19.0 & 21.9 \\ \hline
% 2. & Audio Raw \cite{elizalde2023clap} & 2.97 & 27.5 & 15.3 & 2.1 & 4.6 & 28.7 & 16.7 & 2.6 \\ 
% 3. & Audio-Based & 22.0 & 52.8 & 23.1 & 17.6 & 28.1 & 53.4 & 21.0 & 15.6 \\ \hline
% 4. & Vision-Based &  &  &  &  &  36.5&  58.1&  25.5&  36.2\\ \hline  
% 5. & Fusion & & & & & & & & \\ \hline
% 6. & Full &  &  &  &  & \textbf{43.0}& 61.8& 28.4& 50.5\\ \hline
% \end{tabular}
% \end{table}

\begin{table}[h!]
\centering
\caption{Table result for different models versions on MSR-VTT and VATEX}
\label{tab:merged_result}
\resizebox{\textwidth}{!}{%
\begin{tabular}{|c|l|c|c|c|c|c|c|c|c|c|c|} 
\hline
\multirow{2}{*}{No.} & \multirow{2}{*}{Models} & \multicolumn{5}{|c|}{MSR-VTT} & \multicolumn{5}{|c|}{VATEX} \\   \cline{3-12}
 &  & BLEU@4& ROUGE-L & METEOR & CIDEr & AVG& BLEU@4 & ROUGE-L & METEOR & CIDEr & AVG \\ \hline
1. & BLIP \cite{li2022blip}   & 36.4 & 55.2 & 26.0 & 43.1&40.2 & 21.5& 38.6& 21.3& 29.8&27.8 \\ \hline
2. & Audio Raw \cite{elizalde2023clap}   & 4.6 & 28.7 & 16.7 & 2.6 &13.1 &4.1& 29.5& 10.9&1.2&11.4\\ \hline
3. & Vision-Based & 32.9&  56.4&  25.7&  38.1&38.2   & 23.2& 39.8& 22.1&40.8&31.4\\ \hline  
4. & Audio-Based & 28.1 & 53.4 & 21.0 & 15.6 &29.5 & 19.2& 32.4& 15.5&19.3&21.6\\ \hline
5. & Fusion & 37.2& 59.3& 27.2& 43.0&41.6  & 26.1& 41.1& 22.7& 39.5&32.3\\ \hline
% 6. & MICap &  \textbf{43.0}& 61.8& 28.4& 50.5& \textbf{29.0}& \textbf{44.9}& \textbf{25.1}&\textbf{45.0}\\ \hline
% \end{tabular}
6. & MICap &  \textbf{48.0}& \textbf{63.7}& \textbf{29.4}& \textbf{52.1} & \textbf{48.3} &\textbf{30.5} & \textbf{43.5}& \textbf{23.6}&\textbf{40.9}&\textbf{34.6}\\ \hline
\end{tabular}
}
\end{table}

% \begin{table}
% \centering
% \caption{Table result for different models versions on VATEX}
% \label{tab:result_vatex}
% \begin{tabular}{|l|l|l|l|l|l|} \hline  

% No & Models & BLEU-4 &  ROUGE-L & METEOR & CIDEr \\ \hline  

% 1. & BLIP \cite{li2022blip} & 17.7& 35.2& 17.2& 18.8\\ \hline 
% 2. &  Audio Raw  \cite{elizalde2023clap}& 4.1& 29.5& 10.9&1.2\\\hline \hline 
% 3. &  Audio-Based& 9.7& 32.4& 15.5&4.8\\ \hline 
% 4. &  Vision-Based& 20.0& 42.0& 18&24\\\hline\hline \hline  

% 5. &  Fusion & -& -& -& -\\ \hline
% 6. &  Full & \textbf{29.0}& \textbf{44.9}& \textbf{25.1}&\textbf{45.0}\\\hline 

% \end{tabular}
% \end{table}

% \subsection{State of the Art comparaison}
% \label{sec:sota}

%This ablation study demonstrates the impact of each model component. 

Table \ref{tab:merged_result} shows the result for the MSR-VTT and VATEX datasets for our models variant. In general, we can see that there are gradual improvements on the observed results on both datasets, suggesting the benefit of our approaches. 

Specifically, we first see that the results from the original model that operates in visual domain (BLIP) and Audio domain (Audio Raw) are lower from our own methods (Audio-based / Vision-based). In overall, we see quite an improvement from the results of BLIP to Vision-Based of around 5 scores across metrics (however, we still see high margin on CIDEr score on VATEX around 22). Furthermore,  we see major upgrades of the results from Audio Raw to our Audio-Based method with average improvements of 10 scores accross metrics, showing the benefit of our pipelines.%  the different of the results, when adding different modality the advantage of our processing pipelines. 

We further found that our Fusion approach yields better results in comparison to our methods that operates using only single-modality. We observe about 5 points average improvements between our single-modality based method to our fusion method. This shows the benefits of the the use of multi-modality, given that the model can relate to the importance of each modality, that in our case is done through our attention based fusion mechanism. 

Another finding is that, we observe another gain with the use of our combined loss. Here we see the gains of approximately 6 points across metrics, with highest improvements seen on B@4 and CIDEr (about 6 and 3 of the scores for MSR-VTT and VATEX respectively) achieved by our full method compared to the fusion method. This suggests that the use of contrastive loss, in addition to the standard caption loss, is beneficial in improving the quality of the prediction. This can stem to the additional process that further relates both Visual and Audio modality during image and audio-text alignment process, allowing our method to extract more information from these modalities.

%     we find that adding the fusion mechanism increases accuracy (as seen in the progression from rows 2-4 to 5).
%     \item The addition of contrastive loss results in improved accuracy (evident in the comparison between rows 5 and 6).

% use of image and audio modalities up to our full model. We use BLIP as a base comparaison 

% We observe that our complete model, incorporating all methods, achieves competitive results across all metrics. Specifically, adding the fusion mechanism increases accuracy (as seen in the progression from rows 2-4 to 5). Here we can see that 

% Our full model, which combines all these components, demonstrates the best overall performance.

\subsection{Visual and Audio Interpretability}

\begin{figure*}[!h]
    \centering
    \includegraphics[scale=0.075]{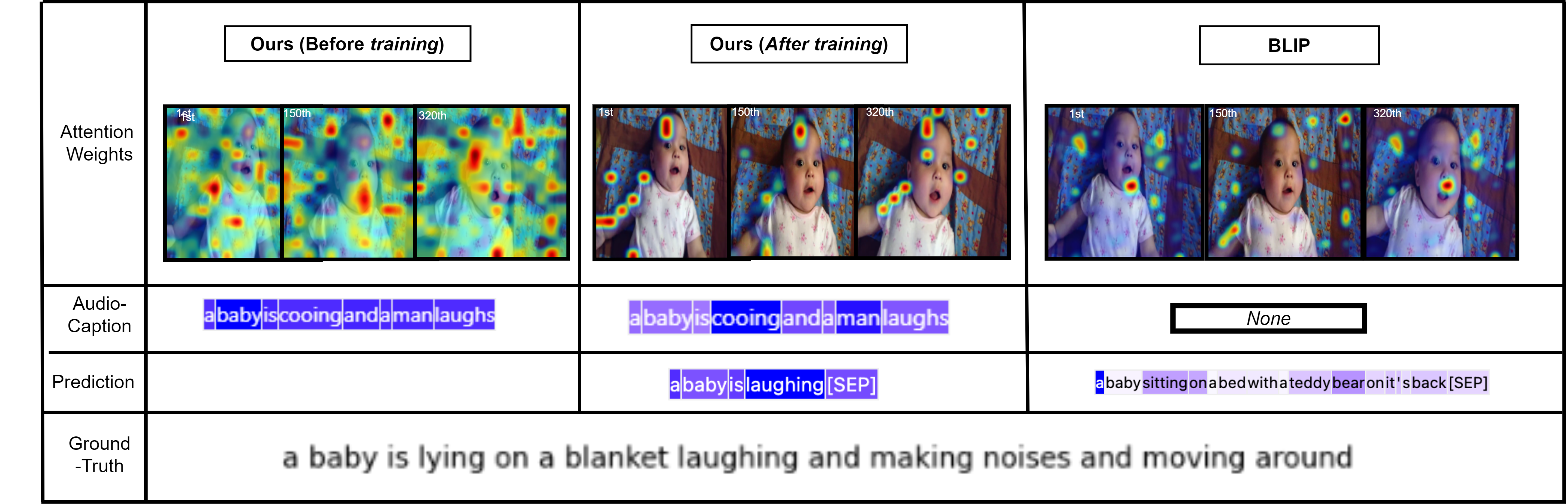}
    \caption[]{Visualisation of different explainability aspect :  Attention weights on the image, audio caption and generated video caption respectively.
    } 
    \label{fig:qual_comp}
\end{figure*}

\label{sec:explain}

To gain insights into the inner workings of our video captioning model, we further analyze our modality processing blocks with respect the their video caption prediction.  There are three targeted blocks for the analysis: First the input video applied to the Video-Audio block with respect to the output (generated caption). Second is the attention mechanisms of audio-captions, applied to the Audio-Video branch in respect to the generated caption. Lastly the self attention during the predictions (our decoder block). This analysis will allow us to inspect where the models attend to during the predictions (with respect to their inputs, the visual and audio modality), thus establishing explanation on the models behavior. 

% This is because, not only BLIP does not process the Audio, but also inherently not capable to explain them 

Fig.~\ref{fig:qual_comp} provides a visualization of the attention mechanisms in our Modalities Combiner employed by our model during the video captioning process. In the figure, we observe that in general, our model provides access to the explanations for predictions in the audio modality compared to other models like BLIP, that lack this capability. 

For example, analyzing attention weights on the visual space of our method reveals stronger activations on the facial area of the target image, which in this case is considered a relevant area in this example. On the other hand, we can see that BLIP instead focuses on sparse and other area, including the background images. Furthermore, our method also provides explanations in the audio space, concentrating on related key-words, such as 'cooing' and 'a man', where indeed there are audio of a baby is cooing and man is talking on the background. This implies that our method is capable to capture the important part of each modalities, related to the event that is happening on the scene.

%Given this, our model demonstrates higher capability to focus on specific areas, while other models tend to have a more generalized focus. 

% Change the trade-off word by something else less strong
%In the same manner as above, in Fig.~\ref{fig:qual_comp} we provide a 

With respect to the visualization of the attention mechanism present in the decoder side to the generated caption, we found that our approach incorporates the relevant audio caption information in relation to the generated caption. For instance, we see that the methods focus "cooing" from the audio modality is highly related to the "laughing", one keyword the model focuses on the generated caption. Although, the attention is not yet perfectly focused, as we still observe some dispersion of attention across less relevant token, however we still see the importance of this keyword in respect to the actual ground truth ("laughing" keyword that is present in the ground truth).

Finally, we observe that there is a compromise in our method: there is trade-off between the precision of the detail of the description of the view with the accuracy of the explanation of the activity of what is happening in the video. However, in this particular case, this might be less relevant considering that the ground truth put more emphasize to the activity related to ongoing events on the video.

\subsection{Explorations of features space on Contrastive loss}
\label{sec:contrastive}

\begin{figure*}[!t]
    \centering
    \includegraphics[scale=0.07]{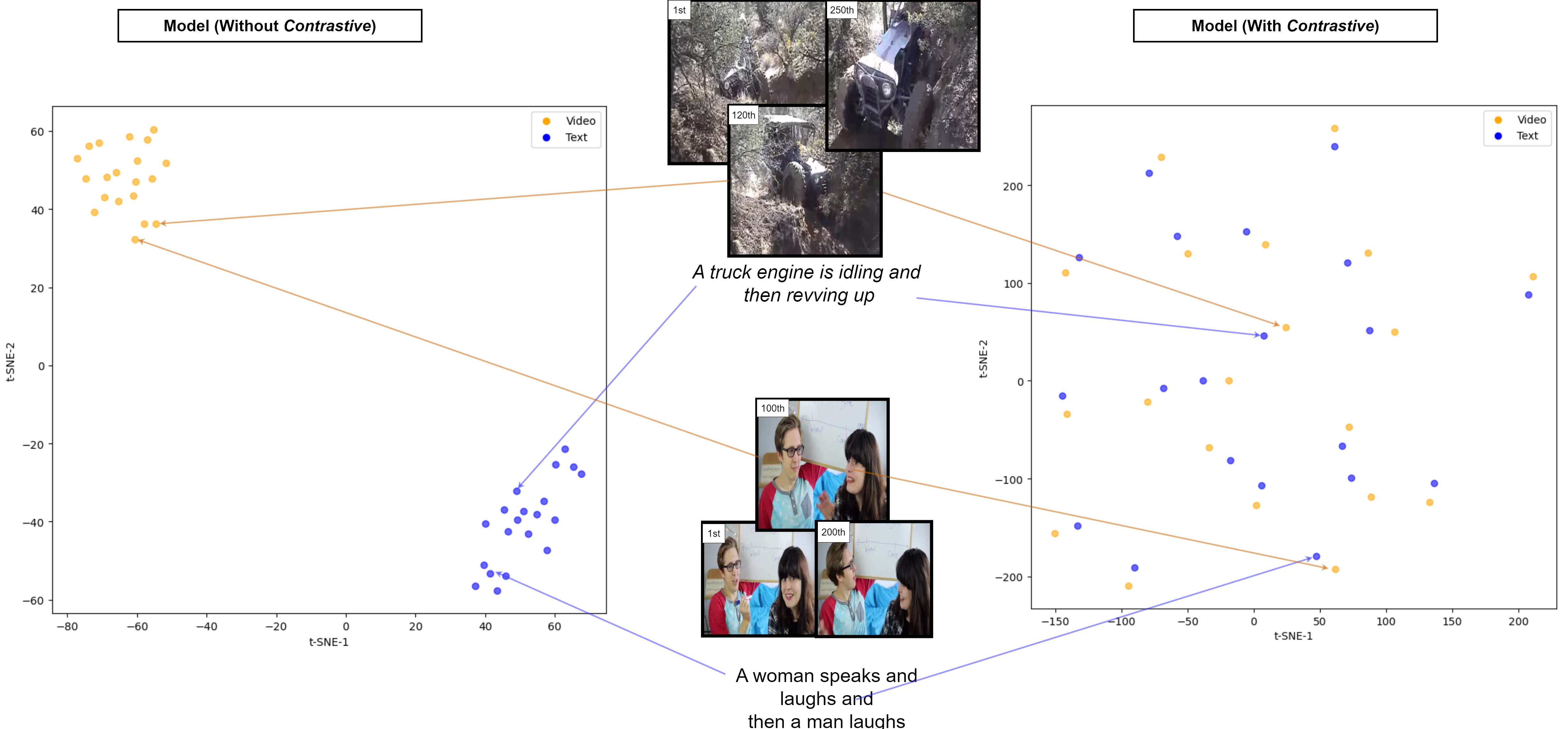}
    \caption[]{Visualization of t-SNE : Fused vs FullModel (MICap)}
    \label{fig:tsne}
\end{figure*}

%452,video7649.mp4,A truck engine is idling and then revving up. 
%24,video7040.mp4,A woman speaks and laughs and then a man laughs. 
%Make it video (multiple image superposed)

Figure ~\ref{fig:tsne} gives us a vizualization of features from our Fusion (without contrastive loss) and Full model after applying t-SNE method between positives and negatives pairs for different features pairs (audio-caption and video frames). Here we can see that the features between the video and audio pair (from the same instance) are located on distinct area. However after the use of contrastive loss (MICap), they are aligned closed together. This observation has been show on other study (e.g. video retrieval), for example BLIP that contrast visual and video caption. However, remarkably this also applies when we substitute the video caption and with the audio caption, with the main difference is that the audio-caption is always accessible during inference (true video caption as input is not present during testing) - that can help in maintaining the quality of the prediction. 

%These representations may helps during learning, as we have seen in improved accuracy on the previous ablation experiment. 

Observing the figure, we can also see that indeed there are implicit differences between pairs of video frames and audio (which suggests that their respective frames and audios should be positioned closed together). For instance, a pair of video frames shows a woman speaking, which closely corresponds to its audio caption mentioning "a woman speaks up" in the background distinct from another pair of video frames example depicting a truck revying up. Modeling these inherent differences, can be considered important during learning and predictions, that might explain the increase on the accuracy obtain when our proposed loss is incorporated.

\subsection{State of the Art Comparisons}
\label{sec:sota}

\begin{table}[ht]
\centering
\caption{Quantitative comparisons on MSR-VTT and VATEX dataset. The best performing results are in bold, the second bests are in red, and the third bests are in blue colour respectively.}
\label{tab:result_sota_msrvtt_vatex}
\resizebox{\textwidth}{!}{%
\begin{tabular}{|l|@{}l|c|c|c|c|c|c|c|c@{}|} 
%\toprule
\hline
\multirow{2}{*}{No.}& \multirow{2}{*}{Models}& \multicolumn{4}{|c|}{MSR-VTT} & \multicolumn{4}{|c|}{VATEX} \\ \cline{3-10}
% \cmidrule(lr){4-6} \cmidrule(lr){7-10}
 & & BLEU@4 & ROUGE-L & METEOR & CIDEr & BLEU@4 & ROUGE-L & METEOR & CIDEr \\ 
\hline %\midrule

1. &SA-LSTM \cite{xu2016msr} & 36.3 & 58.3& 25.5 & 39.9& - & - & - & - \\ \hline 
2. &SAAT \cite{zheng2020syntax}& 41.9& 61.1& 27.7& \textcolor{blue}{51.0}& -& -& -&-\\ \hline 
3. &POS+CG \cite{wang2019controllable} & 42.3& 61.6& \textcolor{blue}{28.2}& 48.7&-&-&-&-\\  \hline
4. &NITS-VC \cite{singh2020nits}& -& -& -& -& 20.0& \textcolor{blue}{42.0}& 18.0&24.0\\ \hline 
5. &SwinBERT \cite{lin2022swinbert} & \textcolor{red}{43.0} & \textcolor{red}{62.2} & \textbf{29.7}& \textbf{55.7} & \textbf{32.3}& \textbf{45.2}& \textbf{26.4}& \textbf{53.9}\\ \hline  
6. &GIT \cite{wang2022git}&32.3& 51.6& 21.4& 37.4& 25.6& 39.6& \textcolor{blue}{21.3}& \textcolor{blue}{33.2}\\ \hline  
7. &BLIP \cite{li2022blip} &36.4	&55.2	&26.0	&43.1  & 21.5& 38.6& 21.3& 29.8 \\ \hline 
8. &MICap (Ours) &  \textbf{48.0}& \textbf{63.7}& \textcolor{red}{29.4}& \textcolor{red}{52.1}& \textcolor{red}{30.5}& \textcolor{red}{43.5}& \textcolor{red}{23.6}& \textcolor{red}{40.9}\\ 
%\bottomrule \hline
\hline
\end{tabular}
}
\end{table}
%%%

%%% IMPOTRANT 
% We use the implementation of the GIT
%There is a change on both the VATEX dataset
%Benchmark VATEX and BLIP 

%Due to a discrepancy between our dataset and the original dataset, and for a fair comparison, we benchmark the models on our reduced datasets. .

Table ~\ref{tab:result_sota_msrvtt_vatex} shows the results of different methods on the test set of MSR-VTT and VATEX dataset. As a note, we have to re-run all evaluated model on both datasets, to take into account the change happening in the dataset (number of videos is reduced), by using the respective implementation that are publicly available. 

%while ranking second or equal on ROUGE-L and METEOR respectively% 

Given the results, we can see that our proposed approach achieves best or second highest points in terms of CIDEr/METEOR and BLEU@4/ROUGE-L respectively for MSR-VTT and VATEX, demonstrating its competitiveness.  Compared to the state of the art model like SwinBERT, our method still compared favorably, with the average margin of less than 5 points accross datasets. However, SwinBERT performance can be attributed from the fact use VideoSwin \cite{liu2022video} as its backbone, which was pre-trained on the Kinetics dataset \cite{kay2017kinetics},  of which the VATEX \cite{wang2019vatex} dataset is a subset. This large amount of dataset used during their training can largely impacts their results. Despite of this, we can still see that our approach achieves superior in two key metrics such as B@4 and Rouge-L on MSR-VTT and VATEX. This can come from the strategic use of attention mechanisms, enhancing the model's ability to focus on relevant features.

\begin{figure*}[!t]
    \centering
    \includegraphics[scale=0.06]{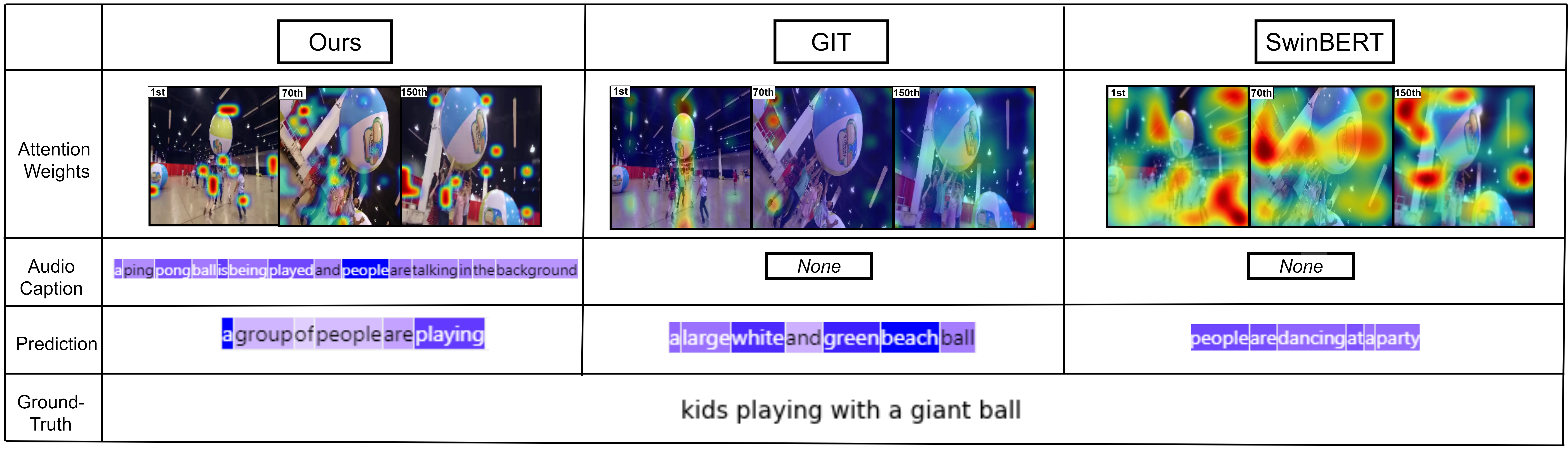}
    \caption[]{Vizualisation of different explainability aspect, on different models :  Attention weights on the image, audio caption and caption prediction respectively 
    } 
    \label{fig:sota_xai}
\end{figure*}

Fig.~\ref{fig:sota_xai} provides comparison of methods focus of our method, GIT and SwinBERT, by computing both GradCAM and evaluating attention weight of each method. In this example, our model focuses on more  elements like bodies, people, and objects (balloons). where the other method lack the focus, as can be seen where GIT, especially SwinBERT attend to almost all region of the input images, including the background. Thus, we can see that our model provides more accessible explanations for predictions in the audio modality compared to other models.  

The improved focus can be attributed to the incorporation of audio modality, which provides us with temporal information for understanding dynamic video content. For example, our model's audio captions offer keywords like "ball," directing visual attention to relevant areas, as evidenced by the model's focus on the ball in both audio and visual domains. Additionally, our model more accurately explains video scenes, such as correctly identifying a group of people playing. 

In terms of the models focus on the generated caption, our model demonstrates capability to focus on specific areas. For instance, in Fig ~\ref{fig:sota_xai} , our model has a string focus on where relevant object  (people and balloon ) are localized.  Moreover, we also provide explanation in the audio space, where it concentrates the most on word like the ball and people (where the other models are lacking). 

Finally, we found that our prediction are more relatable to the ground truth where word like "people" and "playing" are present that describes the object on the videos and also their activity. In other hand, other method such as GIT only explains the object present on the videos. Finally, while SwinBERT also provides both object and the activity, however, they attend largely to all of the words, suggesting lack of the focus in general.

\section{Conclusion}
\label{sec:Conclusion}
In this work, we introduce a novel multi-modal captioning method equipped with interpretability component to address the lack of the utilization of such modalities for video caption generations. Our approach utilizes both of the visual and audio information to benefit from each modalities characteristics. We do this by first encoding both modalities inputs with the use of encoders, followed by the aggregation of these features through our modalities combiner, that leverages on the use of attention mechanisms. Subsequently, we use auto-regressive caption decoder to produce the captions. We further incorporate contrastive loss, during feature representation, to improve the quality of the method by aligning both audio and video to exploit their characteristics of their modality. 

To show the benefit of our approach, we perform multiple comparisons on well known established datasets to show the benefits of our approach progressively throughout of different modalities, fusion mechanisms and further including the contrastive loss, where we also show the capabilities of the method to explain their prediction in both modalities. In our experiment, we show that the model is able to show focus on spatial modality (-video) and text modality (audio-caption). Comparison with other methods shows the competitive results of our approach, that is our method manages to achieve high scores on different quantitative metrics while simultaneously focuses on relevant areas upon each of modalities input.  This in overall demonstrates our high accurate results and explainability capacity of our proposed approach. Future work will focus on scaling our method with larger datasets and finding meaningful feature representations to further improve the quality of the generated captions.
%Combined with fusion block which then follow by a decoder that constrain the generated caption.

\bibliographystyle{splncs04}
\bibliography{sample}

\end{document}